\documentclass{article}
\usepackage{arxiv}
\usepackage[utf8]{inputenc} 
\usepackage[T1]{fontenc}    
\usepackage{url}            
\usepackage{booktabs}       
\usepackage{amsfonts}       
\usepackage{nicefrac}       
\usepackage{microtype}      
\usepackage{lipsum}		
\usepackage{doi}
\usepackage{longtable}
\usepackage{array}

\usepackage{color,array,amsthm}
\usepackage{graphicx}
\usepackage{amsmath,amsthm}
\usepackage{epstopdf}
\usepackage{lineno,hyperref}
\usepackage{longtable,tabularx}
\usepackage[version=4]{mhchem}
\usepackage{siunitx}
\usepackage{float}
\usepackage{dsfont}

\usepackage{amssymb}
\usepackage{array}
\usepackage[T1]{fontenc}
\usepackage{tabularx}
\usepackage{color}
\usepackage{mathtools}
\usepackage{cancel}
\usepackage{subfig}
\usepackage{xcolor}
\usepackage[linesnumbered,lined,boxed,commentsnumbered]{algorithm2e}
\usepackage{bm}
\newcommand{\bs}{\bm}
\usepackage{epigraph}

\usepackage{tikz}
\usepackage{tikz-3dplot}
\usetikzlibrary{calc,positioning}
\newcommand{\mat}[1]{\mathbf{#1}}

\theoremstyle{theorem}
\newtheorem{remark}{Remark}

\title{Tensor Invariant Data-Assisted Control and Dynamic Decomposition of Multibody Systems}

\author{
  Mostafa Eslami, Maryam Babazadeh \\
  Department of Mechanical and Process
Engineering\\
RPTU Kaiserslautern-Landau\\
67663 Kaiserslautern, Germany,\\
\tt\small 
  \texttt{\{m.eslami, m.babazadeh\}@rptu.de} 
}

\renewcommand{\shorttitle}{Tensor Invariant Data-Assisted Control}

\begin{document}
\maketitle

\begin{abstract}
The control of robotic systems in complex, shared collaborative workspaces presents significant challenges in achieving robust performance and safety when learning from experienced or simulated data is employed in the pipeline. A primary bottleneck is the reliance on coordinate-dependent models, which leads to profound data inefficiency by failing to generalize physical interactions across different frames of reference. This forces learning algorithms to rediscover fundamental physical principles in every new orientation, artificially inflating the complexity of the learning task.
This paper introduces a novel framework that synergizes a coordinate-free, unreduced multibody dynamics and kinematics model based on tensor mechanics with a Data-Assisted Control (DAC) architecture. A non-recursive, closed-form Newton-Euler model in an augmented matrix form is derived that is optimized for tensor-based control design. This structure enables a principled decomposition of the system into a structurally certain, physically grounded part and an uncertain, empirical, and interaction-focused part, mediated by a virtual port variable. Then, a complete, end-to-end tensor-invariant pipeline for modeling, control, and learning is proposed.
The coordinate-free control laws for the structurally certain part provide a stable and abstract command interface, proven via Lyapunov analysis. Eventually, the model and closed-loop system are validated through simulations. This work provides a naturally ideal input for data-efficient, frame-invariant learning algorithms, such as equivariant learning, designed to learn the uncertain interaction. The synergy directly addresses the data-inefficiency problem, increases explainability and interpretability, and paves the way for more robust and generalizable robotic control in interactive environments.
\end{abstract}

\keywords{Tensor Mechanics \and Multibody Systems \and Robotics \and Data-Assisted Control \and Dynamic Decomposition \and Equivariant Learning}

\section{Introduction}
\label{sec:intro}

The next frontier in robotics involves moving from structured industrial settings to complex, dynamic, and human-centric environments. Achieving this vision requires control systems that are not only precise and robust but also capable of safe, predictive interaction and continuous improvement from experience. These multifaceted challenges demand a holistic approach that integrates the predictive power of data with the reliability of physical models. Data-Assisted Control (DAC) has emerged as a powerful paradigm for managing uncertainty in complex systems \cite{eslami2023sequential,eslami2024data,eslami2025generalization,eslami2025learning}. It operates on the principle of decomposing the system dynamics into a well-modeled, structurally certain Left-Hand Side (LHS) and an uncertain, often algebraic, Right-Hand Side (RHS). These are connected by a virtual port variable, which serves as the interface between a model-based controller and a data-driven component. However, the efficacy of any DAC architecture is fundamentally constrained by the representation used for the known LHS dynamics. If the physical model of a multibody system is expressed in a coordinate-dependent manner, the learning task for the RHS becomes unnecessarily complex and data-intensive, as it must implicitly learn to disentangle physical principles from representational artifacts.

For robots interacting with the physical world, frame-invariance is a critical requirement for data efficiency. As an example, the same external force applied in two different geographical locations has the same physical effect, yet it would be interpreted as novel data by a learning algorithm tied to a specific coordinate system. This necessitates a framework where physical models, control, and learning are all expressed in a coordinate-free manner. Equivariant neural networks provide a powerful tool for this by building symmetries directly into the model architecture, leading to dramatic improvements in sample efficiency and generalization \cite{cohen2016group,cohen2019general,smets2024mathematicsneuralnetworkslecture}. Recent research in robotics has increasingly leveraged equivariant representations for tasks from perception to control. The framework proposed herein achieves a unique synergy: the tensor mechanics model provides a naturally coordinate-free representation of the system's physical state (e.g., velocities, forces as geometric objects), which serves as the ideal input for an equivariant learning algorithm designed to learn the uncertain interaction dynamics (RHS). This creates a complete, end-to-end frame-invariant pipeline.

To realize this coordinate-free vision, instead of vector mechanics, a tensor mechanics within the Newton-Euler formalism is followed, inspired by its success in aerospace applications \cite{zipfel2014modeling}. This allows to derive a non-recursive, unreduced, closed-form model of the multibody dynamics and kinematics, that is distinguished from its recursive counterpart \cite{zipfeltensorial}. The choice of an unreduced, non-recursive formulation, which results in a set of Differential-Algebraic Equations (DAEs) in an augmented matrix form, is a deliberate design choice optimized for control, not just simulation. An unreduced model simplifies the incorporation of external interactions. A non-recursive, augmented form explicitly solves for internal constraint forces, which are crucial for the DAC decomposition and the subsequent Lyapunov stability analysis. This structure, while potentially less computationally efficient than recursive methods for large-scale simulation, is structurally superior for designing and proving the stability of the proposed controller.

This work is situated within the broader field of geometric mechanics, yet it offers a distinct approach. The dominant paradigm for coordinate-free robot dynamics is geometric mechanics on Lie groups, where configurations are represented on the manifold $SE(3)$. Seminal works established elegant formulations that often lead to highly efficient $O(n)$ recursive algorithms for forward and inverse dynamics \cite{park1995lie,polen1997lie,polen1999coordinate}. While both tensor-based approach and Lie group methods achieve coordinate-invariance, they lead to fundamentally different algorithmic structures. Lie group methods typically yield recursive algorithms that are optimized for simulation by eliminating internal constraint forces. In contrast, a non-recursive, augmented DAE formulation is optimized for control design by explicitly exposing these constraint forces as Lagrange multipliers, which are essential variables within our DAC framework and its stability analysis.

This approach also contrasts with purely data-driven methods for modeling system dynamics, such as Neural Ordinary Differential Equations (NODEs), which learn the derivative of the system's state from data \cite{duong2024port}. Unlike NODEs, which treat the entire dynamics as a black box to be learned, the DAC framework is a physics-informed or gray-box approach. By encoding the known, certain physics (mass, inertia, kinematics) in the LHS, it simplifies the learning problem for the RHS. This leads to greater data efficiency, interpretability, and stronger, structure-based stability guarantees for the core system dynamics compared to proving stability for a learned black-box model.

To provide further context for the modeling choices, the landscape of multibody dynamics formulations for control applications shows that the Newton-Euler method offers a direct and computationally efficient path to the equations of motion, often preferred over the energy-based Lagrangian approach for its clarity in representing forces and moments. Formulations can be recursive, offering $O(n)$ complexity ideal for large-scale simulation \cite{featherstone2014rigid,bae2001generalized}, or non-recursive and closed-form, which provides greater structural transparency for control design. While implicit formulations can offer superior numerical stability \cite{Boyer2024}, explicit closed-form models are of greater interest for high-bandwidth control applications due to their computational speed and more predictable uncertainty bounds \cite{bascetta2017}. For systems with flexible components, methods range from computationally intensive Finite Element Methods (FEM) \cite{shabana1997flexible,nandihal2022dynamics} to specialized models like Cosserat rods \cite{Boyer2024,Renda2018}, but these often introduce complexities that are challenging for real-time control. Our work focuses on rigid-body systems with joint flexibility, for which a closed-form, explicit Newton-Euler model provides an optimal balance of fidelity, computational tractability, and structural suitability for our proposed control architecture.

In previous works, the DAC framework was first established for flight dynamics, where the system model was partitioned into its certain (internal) and uncertain (external) parts. A model-based nonlinear controller handled the known internal dynamics, while data-driven Koopman operators were used to learn the uncertain external aerodynamic forces and moments from pseudo-observations~\cite{eslami2023sequential,eslami2024data}. To broaden this concept, the framework was then generalized for any system that can be described using port-Hamiltonian (pH) mechanics~\cite{eslami2025generalization}. This work introduced a principled decomposition of the dynamics into a conservative, model-based LHS  and a Reinforcement Learning agent for the uncertain dissipative/input-driven RHS, connected by a virtual port variable. Here, the RL agent's objective is to learn a policy that maps the desired port command from the LHS to the physical control inputs, thereby maximizing a reward function aligned with the LHS performance objectives. This generalization was framed around a set of core hypotheses regarding the complexity of the learning task, the guaranteed existence of a stable LHS controller under varied uncertainty and observability, the ability of the RL agent to satisfy safety constraints, and the synergistic link between Persistency of Excitation (PE) for LHS estimation and the sample complexity of RHS learning. 

However, applying this architecture to systems with fast dynamics, such as the highly complex model of a free-floating space manipulator, revealed a critical challenge. A direct implementation of a Deep Neural Network (DNN) on the RHS produced poor performance, as the potential mismatch between the desired port variable and the learned value risked instability~\cite{eslami2025learning}. This issue was resolved by introducing an intermediate high-gain classical controller on the RHS, which created a strict time-scale separation that successfully stabilized the DNN. While effective, this solution highlighted the necessity for improved sample efficiency and a more fundamental modeling paradigm capable of handling the intricate dynamics of multi-robot collaboration, where interaction forces change the system topology.

This paper establishes that necessary foundation. We propose employing coordinate-free tensor mechanics to formulate both the multibody dynamics and the control laws. This approach yields an invariant model that is inherently robust to the choice of coordinate systems and provides a pragmatic, structured basis for handling the complexities of physical interaction. By first delivering a rigorous, tensor-based model and control design for a general multibody system, this work paves the way for future investigations into complex, multi-agent DAC architectures, whether centralized or decentralized. In summary, the main contributions of this paper are:

\begin{enumerate}
    \item A novel coordinate-free dynamics framework for multibody systems using tensor mechanics, resulting in an unreduced, non-recursive, closed-form model suitable for control.
    \item The principled application of the DAC paradigm to this tensor-based model, creating a clean and structurally advantageous separation between certain physical dynamics and uncertain interactions.
    \item The design of coordinate-free, tensor-based control laws with rigorous Lyapunov stability proofs, providing a stable and abstract command interface for the uncertain system components.
    \item A clear architectural pathway for integrating data-efficient learning methods, specifically equivariant learning, by creating a fully frame-invariant modeling, control, and learning pipeline.
\end{enumerate}
The remainder of this paper is organized as follows. Section \ref{sec:modeling} details the derivation of the tensor-based multibody dynamics model. Section \ref{sec:control} presents the data-assisted control design and stability analysis. Section \ref{sec:simulation} validates the framework through numerical simulation, and Section \ref{sec:conclusion} provides concluding remarks.

\section*{Nomenclature}
\renewcommand{\arraystretch}{1.1}
\begin{longtable}{>{\(}l<{\)} p{0.75\textwidth}}
\caption{Nomenclature of variables and symbols used throughout the paper.} \label{tab:nomenclature} \\
\toprule
\textbf{Symbol} & \textbf{Description} \\
\midrule
\endfirsthead

\multicolumn{2}{c}%
{{\bfseries \tablename\ \thetable{} -- continued from previous page}} \\
\toprule
\textbf{Symbol} & \textbf{Description} \\
\midrule
\endhead

\midrule
\multicolumn{2}{r}{{Continued on next page}} \\
\endfoot

\bottomrule
\endlastfoot

\multicolumn{2}{l}{\textbf{Indices and General Symbols}} \\
N & Number of rigid bodies in the system, $\mathbb{N}$. \\
i & Index for a rigid body, $i \in \{1, \dots, N\}$. \\
j & Index for a joint, $j \in \{1, \dots, N-1\}$. \\
I & Inertial reference frame. \\
B_i & Body-fixed reference frame for body $i$. \\
J_j & Location of joint $j$. \\
\delta_{ij} & Kronecker delta function. \\

\addlinespace[1em]
\multicolumn{2}{l}{\textbf{Operators and Transformations}} \\
D^I(\cdot) & Rotational time derivative operator with respect to Inertial frame. \\
\mat{T}^{B_iI} & Transformation tensor from frame $I$ to frame $B_i$, $SO(3)$. \\
{[\bs{a}]^X} & Coordinate representation of tensor $\bs{a}$ in frame $X$. \\

\addlinespace[1em]
\multicolumn{2}{l}{\textbf{Kinematic Tensors}} \\
\bs{s}_{XY} & Position tensor from point Y to point X, $\mathbb{R}^{3 \times 1}$ [m]. \\
\mat{S}_{XY} & Skew-symmetric tensor corresponding to the position tensor $\bs{s}_{XY}$, $\mathbb{R}^{3 \times 3}$. \\
\bs{v}_{B_i}^I & Linear velocity tensor of the center of mass of body $i$ in frame $I$, $\mathbb{R}^{3 \times 1}$ [m/s]. \\
\bs{\omega}_{B_i}^I & Angular velocity tensor of body $i$ in frame $I$, $\mathbb{R}^{3 \times 1}$ [rad/s]. \\
\bs{\Omega}_{B_i}^I & Skew-symmetric tensor of the angular velocity tensor $\bs{\omega}_{B_i}^I$, $\mathbb{R}^{3 \times 3}$. \\
\bs{\nu}_v & Stack of all body linear velocity tensors, $\mathbb{R}^{3N \times 1}$. \\
\bs{\nu}_\omega & Stack  of all body angular velocity tensors, $\mathbb{R}^{3N \times 1}$. \\
\bs{\nu} & Generalized velocity tensor (twist), $\mathbb{R}^{6N \times 1}$. \\
\bs{\gamma}_j & Velocity-product acceleration tensor at joint $j$, $\mathbb{R}^{3 \times 1}$ [m/s$^2$]. \\
\bs{\gamma} & Stack of all velocity-product tensors, $\mathbb{R}^{3(N-1) \times 1}$. \\
\mat{J} & System constraint Jacobian tensor, $\mathbb{R}^{3(N-1) \times 6N}$. \\

\addlinespace[1em]
\multicolumn{2}{l}{\textbf{Dynamic and Force Tensors}} \\
m_{B_i} & Mass of body $i$, $\mathbb{R}^+$ [kg]. \\
\mat{I}_{B_i} & Inertia tensor of body $i$, $\mathbb{R}^{3 \times 3}$ [kg$\cdot$m$^2$]. \\
\bs{g} & Gravitational acceleration tensor, $\mathbb{R}^{3 \times 1}$ [m/s$^2$]. \\
\bs{f}_{B_i} & External force tensor applied to body $i$, $\mathbb{R}^{3 \times 1}$ [N]. \\
\bs{m}_{B_i} & External moment tensor applied to body $i$, $\mathbb{R}^{3 \times 1}$ [N$\cdot$m]. \\
\bs{f}_{J_j} & Internal constraint force tensor at joint $j$, $\mathbb{R}^{3 \times 1}$ [N]. \\
\bs{m}_{J_j} & Internal moment tensor at joint $j$, $\mathbb{R}^{3 \times 1}$ [N$\cdot$m]. \\
\mat{F}_B & Stack of external force tensors, $\mathbb{R}^{3N \times 1}$. \\
\mat{M}_B & Stack of external moment tensors, $\mathbb{R}^{3N \times 1}$. \\
\mat{F}_J & Stack of internal joint force tensors (Lagrange multipliers), $\mathbb{R}^{3(N-1) \times 1}$. \\
\mat{M}_J & Stack of internal joint moment tensors, $\mathbb{R}^{3(N-1) \times 1}$. \\
\mat{M}_v & System mass tensor for linear dynamics, $\mathbb{R}^{3N \times 3N}$. \\
\mat{M}_\omega & System inertia tensor for angular dynamics, $\mathbb{R}^{3N \times 3N}$. \\
\mathcal{M} & Generalized mass tensor, $\mathbb{R}^{6N \times 6N}$. \\
\bs{\mathcal{C}}(\bs{\nu}) & Coriolis and gyroscopic effects tensor, $\mathbb{R}^{6N \times 6N}$. \\
\mathcal{F} & Generalized applied force tensor, $\mathbb{R}^{6N \times 1}$. \\

\addlinespace[1em]
\multicolumn{2}{l}{\textbf{Control Tensors and Variables}} \\
\bs{\tau} & Virtual port tensor (generalized interaction force/moment), $\mathbb{R}^{6N \times 1}$. \\
\bs{\tau}_c & Commanded port tensor, $\mathbb{R}^{6N \times 1}$. \\
\bs{\nu}_d & Desired generalized velocity trajectory, $\mathbb{R}^{6N \times 1}$. \\
\bs{s} & Sliding surface error tensor, $\mathbb{R}^{6N \times 1}$. \\
\mat{K}_d & Positive-definite controller gain tensor, $\mathbb{R}^{6N \times 6N}$. \\
V & Lyapunov candidate function, $\mathbb{R}$. \\
\bs{e}_p, \bs{e}_v & Position and velocity error tensors, $\mathbb{R}^{6N \times 1}$. \\
\mat{\Lambda} & Positive-definite gain tensor for composite error, $\mathbb{R}^{6N \times 6N}$. \\
\bs{\nu}_r & Reference velocity tensor, $\mathbb{R}^{6N \times 1}$. \\
\mat{J}_t & Task-space Jacobian tensor for the end-effector, $\mathbb{R}^{6 \times 6N}$. \\
\mat{P} & Null-space projection tensor, $\mathbb{R}^{6N \times 6N}$. \\

\addlinespace[1em]
\multicolumn{2}{l}{\textbf{Attitude Representation}} \\
(\phi_j, \theta_j, \psi_j) & Euler angles for relative orientation at joint $j$ [rad]. \\
\bs{q}^{B_iI} & Quaternion representing orientation of frame $B_i$ w.r.t. $I$, $\mathbb{H}$ (the set of quaternions). \\

\end{longtable}

\section{Modeling}\label{sec:modeling}
\begin{figure}
    \centering
    \includegraphics[width=0.7\linewidth]{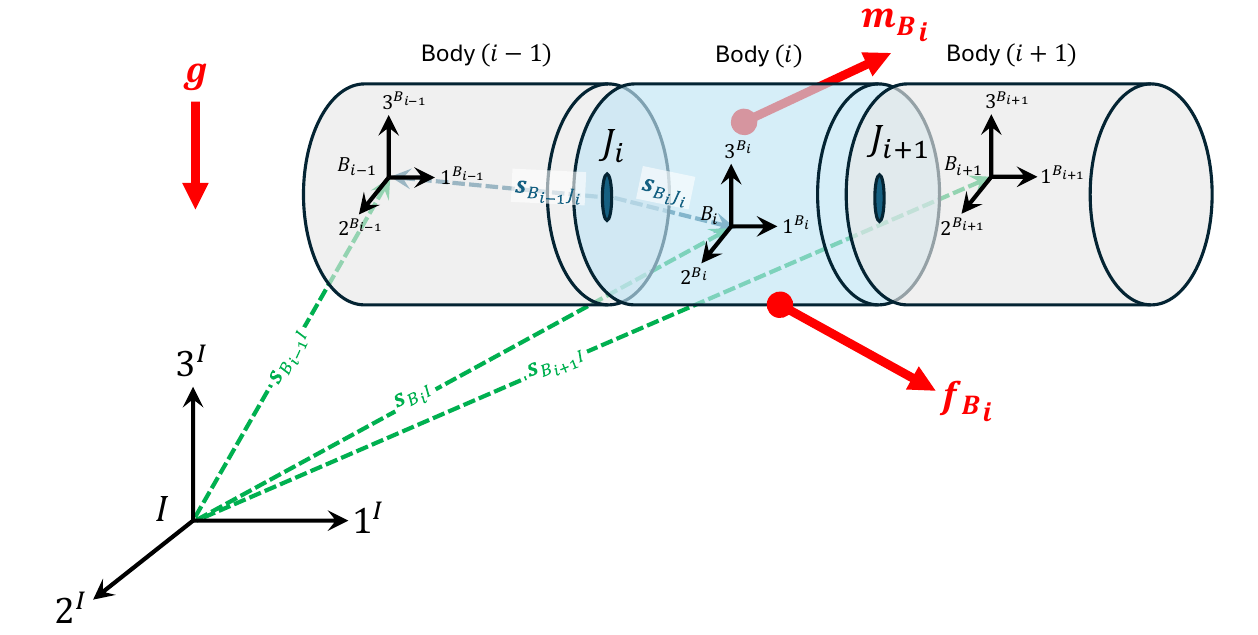}
    \caption{Multibody system diagram}
    \label{fig:multibody}
\end{figure}

The multibody system under study consists of $N$ rigid bodies, connected in a serial chain by $N-1$ joints. The bodies are indexed by $i = 1, 2, \ldots, N$, and the joints are indexed by $j = 1, 2, \ldots, N-1$. Joint $j$ connects body $j$ to body $j+1$. Figure~\ref{fig:multibody} provides a general overview of the multibody system under study with major tensors involved in dynamics. The external force $\bs{f}_{B_i}$ also exerts an associate moment that is combined into $\bs{m}_{B_i}$ tensor. Body frames are located at the center of mass of each body. For position an angular velocity tensors, the corresponding skew-symmetric matrix is denoted by a capital letter, e.g., $\mat{S}$ for a position tensor $\bs{s}$ and $\bs{\Omega}$ for an angular velocity tensor $\bs{\omega}$, such that $\mat{S}\bs{\omega} = \bs{s} \times \bs{\omega} = -\bs{\Omega}\bs{s}$.

\subsection{Linear Momentum}
Following Newton's second law, the equations for the linear momentum of each body are formulated. We account for external forces $\bs{f}_{B_i}$, gravitational forces $m_{B_i}\bs{g}$, and internal joint reaction forces. The reaction force exerted by body $j$ onto body $j+1$ at joint $j$ is denoted by $\bs{f}_{J_j}$. The rotational time derivative of the linear momentum $\bs{p}_{B_i}^I = m_{B_i}\bs{v}_{B_i}^I$ for each body $i$ is given by the sum of forces acting on it:
\begin{align}
\left\{
\begin{array}{l}
    m_{B_1} D^I\bs{v}_{B_1}^I = \bs{f}_{B_1} + \bs{f}_{J_1} + m_{B_1}\bs{g} \\
    \quad\vdots \\
    m_{B_k} D^I\bs{v}_{B_k}^I = \bs{f}_{B_k} - \bs{f}_{J_{k-1}} + \bs{f}_{J_k} + m_{B_k}\bs{g}, \quad k \in [2, N-1] \\
    \quad\vdots \\
    m_{B_N} D^I\bs{v}_{B_N}^I = \bs{f}_{B_N} - \bs{f}_{J_{N-1}} + m_{B_N}\bs{g}
\end{array}
\right.
\end{align}
This system of $N$ tensor equations can be consolidated into a single matrix equation. Let us define the stacked state and force tensors: $\bs{\nu}_v = [\bs{v}_{B_1}^{I\intercal}, \ldots, \bs{v}_{B_N}^{I\intercal}]^\intercal \in \mathbb{R}^{3N}$ is the stacked tensor of linear velocities,  $\mat{F}_B = [\bs{f}_{B_1}^\intercal, \ldots, \bs{f}_{B_N}^\intercal]^\intercal \in \mathbb{R}^{3N}$ is the stacked tensor of external forces, and $\mat{F}_J = [\bs{f}_{J_1}^\intercal, \ldots, \bs{f}_{J_{N-1}}^\intercal]^\intercal \in \mathbb{R}^{3(N-1)}$ is the stacked tensor of internal joint forces. The equations of motion in matrix form will be:
\begin{align}
\mat{M}_v D^I\bs{\nu}_v = \mat{F}_B + \mat{C}_F \mat{F}_J + \mat{M}_v \mat{g}_{3N}    
\end{align}
where the system mass tensor $\mat{M}_v = \operatorname{blockdiag}(m_{B_1}\mat{I}_3, \ldots, m_{B_N}\mat{I}_3) \in \mathbb{R}^{3N \times 3N}$, with $\mat{I}_3$ being the $3 \times 3$ identity tensor, the gravitational tensor $\mat{g}_{3N} = [\bs{g}^\intercal, \ldots, \bs{g}^\intercal]^\intercal \in \mathbb{R}^{3N}$, and the force distribution matrix $\mat{C}_F \in \mathbb{R}^{3N \times 3(N-1)}$ maps joint forces to the bodies as follows:
\begin{align}
    \mat{C}_F = \begin{bmatrix}
\mat{I}_3 & \mat{0} & \cdots & \mat{0} \\
-\mat{I}_3 & \mat{I}_3 & & \vdots \\
\mat{0} & \ddots & \ddots & \mat{0} \\
\vdots & & -\mat{I}_3 & \mat{I}_3 \\
\mat{0} & \cdots & \mat{0} & -\mat{I}_3
\end{bmatrix}
\end{align}

\subsection{Angular Momentum}
The rotational dynamics are described by Euler's equations. The rate of change of angular momentum $\bs{l}_{B_i}^I = \mat{I}_{B_i}\bs{\omega}_{B_i}^I$ with respect to the inertial frame and about the center of mass of body $i$ equals the sum of applied moments, i.e.
\begin{align}
    D^I\bs{l}_{B_i}^I = \mat{I}_{B_i}D^I\bs{\omega}_{B_i}^I + \bs{\Omega}_{B_i}^I \mat{I}_{B_i}\bs{\omega}_{B_i}^I = \sum \bs{m}_{B_i}.
\end{align}
The total moment on body $i$ arises from external moment tensor $\bs{m}_{B_i}$, net joint moment tensor $\bs{m}_{J_{i}} - \bs{m}_{J_{i-1}}$, and moment from joint forces $\mat{S}_{B_i J_{i-1}}\bs{f}_{J_{i-1}} - \mat{S}_{B_i J_i}\bs{f}_{J_i}$. Then, the set of angular momentum equations is written as:
\begin{align}
   \left\{
\begin{array}{l}
\mat{I}_{B_1}D^I\bs{\omega}_{B_1}^I = \bs{m}_{B_1} + \bs{m}_{J_1} - \mat{S}_{B_1J_1}\bs{f}_{J_1} - \bs{\Omega}_{B_1}^I\mat{I}_{B_1}\bs{\omega}_{B_1}^I \\[1ex]
\quad\vdots \\[1ex]
\mat{I}_{B_k}D^I\bs{\omega}_{B_k}^I = \bs{m}_{B_k} + \bs{m}_{J_k}- \bs{m}_{J_{k-1}}  + \mat{S}_{B_k J_{k-1}}\bs{f}_{J_{k-1}} - \mat{S}_{B_k J_k}\bs{f}_{J_k} - \bs{\Omega}_{B_k}^I\mat{I}_{B_k}\bs{\omega}_{B_k}^I \quad k \in [2, N-1]\\[1ex]
\quad\vdots \\[1ex]
\mat{I}_{B_N}D^I\bs{\omega}_{B_N}^I = \bs{m}_{B_N} - \bs{m}_{J_{N-1}} + \mat{S}_{B_N J_{N-1}}\bs{f}_{J_{N-1}} - \bs{\Omega}_{B_N}^I\mat{I}_{B_N}\bs{\omega}_{B_N}^I
\end{array}.
\right. 
\end{align}
It can also be written in a compact matrix form. Let us define the stacked tensor of angular velocities: $\bs{\nu}_\omega = [\bs{\omega}_{B_1}^{I\intercal}, \ldots, \bs{\omega}_{B_N}^{I\intercal}]^\intercal \in \mathbb{R}^{3N}$, the stacked tensor of external moments: $\mat{M}_B = [\bs{m}_{B_1}^\intercal, \ldots, \bs{m}_{B_N}^\intercal]^\intercal \in \mathbb{R}^{3N}$, the stacked tensor of internal joint moments: $\mat{M}_J = [\bs{m}_{J_1}^\intercal, \ldots, \bs{m}_{J_{N-1}}^\intercal]^\intercal \in \mathbb{R}^{3(N-1)}$, and the stacked tensor of gyroscopic moments: $\bs{\tau}_{gyro} = [(\bs{\Omega}_{B_1}^I\mat{I}_{B_1}\bs{\omega}_{B_1}^I)^\intercal, \ldots, (\bs{\Omega}_{B_N}^I\mat{I}_{B_N}\bs{\omega}_{B_N}^I)^\intercal]^\intercal \in \mathbb{R}^{3N}$. A matrix form of this term is defined such that it reproduces the gyroscopic torques when multiplied by the generalized velocity. The matrix $\bs{\mathcal{C}}(\bs{\nu})$ is non-zero only in its bottom-right, angular-angular quadrant:
\begin{align}
    \bs{\mathcal{C}}(\bs{\nu}) = 
\begin{bmatrix}
\mat{0} & \mat{0} \\
\mat{0} & \bs{\mathcal{C}}_{\omega\omega}
\end{bmatrix}
\quad \text{where} \quad
\bs{\mathcal{C}}_{\omega\omega} = \operatorname{blockdiag}\left(\bs{\Omega}_{B_1}^I\mat{I}_{B_1}^I,\ldots,\bs{\Omega}_{B_N}^I\mat{I}_{B_N}^I\right) .
\end{align}
Then, the matrix form of the angular momentum equations is written as:
\begin{align}
    \mat{M}_\omega D^I\bs{\nu}_\omega = \mat{M}_B + \mat{C}_F \mat{M}_J + \mat{C}_M \mat{F}_J - \bs{\mathcal{C}}_{\omega\omega}\bs{\nu}_\omega
\end{align}
where the inertia matrix $\mat{M}_\omega = \operatorname{blockdiag}(\mat{I}_{B_1}, \ldots, \mat{I}_{B_N}) \in \mathbb{R}^{3N \times 3N}$, and the moment distribution matrix $\mat{C}_M \in \mathbb{R}^{3N \times 3(N-1)}$ maps joint forces to moments about each body's center of mass. Its block entries are given by $(\mat{C}_M)_{k,j} = \delta_{k,j+1}\mat{S}_{B_k J_j} - \delta_{k,j}\mat{S}_{B_k J_j}$, where $\delta$ is the Kronecker delta. The matrix has the explicit block structure:
\begin{align}
   \mat{C}_M = \begin{bmatrix}
-\mat{S}_{B_1 J_1} & \mat{0} & \cdots & \mat{0} \\
\mat{S}_{B_2 J_1} & -\mat{S}_{B_2 J_2} & & \vdots \\
\mat{0} & \ddots & \ddots & \mat{0} \\
\vdots & & \mat{S}_{B_{N-1} J_{N-2}} & -\mat{S}_{B_{N-1} J_{N-1}} \\
\mat{0} & \cdots & \mat{0} & \mat{S}_{B_N J_{N-1}}
\end{bmatrix}.
\end{align}
The generated internal moment is given by,
\begin{align}\label{equ:ang_con1}
  \bs{m}_{J_j}=K_{\phi_j}\phi_j \bs{j}_{j_1}+K_{\theta_j}\theta_j \bs{j}_{j_2}+K_{\psi_j}\psi_j \bs{j}_{j_3}
\end{align}
to model a flexible joint, and triad $(\phi_j,\theta_j,\psi_j)$ defined as Euler angles of body $j+1$ with respect to body $j$, and triad $(\bs{j}_{j_1},\bs{j}_{j_2},\bs{j}_{j_3})$ is axis of rotation for joint $j$. The rotation axis can be calculated by the Euler angles principal rotation tensor. The transformation matrix of body $j$ with respect to body $j+1$ is given by,
\begin{align}
  ]^{B_j}\xleftarrow[1]{\phi_j}]^{Y_j}\xleftarrow[2]{\theta_j}]^{X_j}\xleftarrow[3]{\psi_j}]^{B_{j+1}} .
\end{align}
Hence, the joint $j$ rotation axis is equivalent to $(\bs{b}_{j_1},\bs{x}_{j_2},\bs{b}_{{j+1}_3})$. The basis $\bs{b}_{j_1}$ and $\bs{b}_{{j+1}_3}$ are directly available in inertial coordinate from first and third rows of their bodies transforation matrix with respect to inertial frame, i.e. $[\mat{T}]^{B_jI}$ and $[\mat{T}]^{B_{j+1}I}$ respectively. The basis $\bs{x}_{j_2}$ is second row of $[\mat{T}]^{X_jB_{j+1}}$. Consequently, it is the second row of the following transformation matrix in the inertial coordinate,
\begin{align}
  [\mat{T}]^{X_jI} = \left[\begin{array}{ccc}
          C_{\psi_j} & S_{\psi_j} & 0 \\
          -S_{\psi_j} & C_{\psi_j} & 0 \\
          0 & 0 & 1
        \end{array}\right][\mat{T}]^{B_{j+1}I}
\end{align}
where $\cos (\alpha)\doteq C_\alpha$, $\sin(\alpha)\doteq S_\alpha$. In this formulation, the internal joint moment tensor, $\mat{M_J}$, is determined by a constitutive law, meaning it is treated as a known function of the system's state (e.g., as a torsional spring dependent on the relative angles between adjacent bodies). If this were not the case (for instance, if a joint's motion were locked), these moments would become unknown constraint moments. As such, they would be solved for as part of the constraint equations, similar to the internal forces in $\mat{F_J}$. By defining $\mat{M_J}$ as a known function, this assumption simplifies the model but does not affect the generality of the subsequent control law, which acts on the independent port variable.

\subsection{Joint Kinematics}
The connectivity of the chain is enforced by kinematic constraints at each joint, i.e.
\begin{align}
\bs{s}_{B_jJ_j}+\bs{s}_{J_jB_{j+1}} = \bs{s}_{B_jB_{j+1}} = \bs{s}_{B_jI} - \bs{s}_{B_{j+1}I} .
\end{align}
Taking the rotational derivative with respect to the inertial frame leads to,
\begin{align}
    D^I \bs{s}_{B_jI} - D^I \bs{s}_{B_{j+1}I} = \underline{D^{B_j}\bs{s}_{B_jJ_j}} + \bs{\Omega}_{B_j}^I\bs{s}_{B_jJ_j} - \underline{D^{B_{j+1}}\bs{s}_{B_{j+1}J_j}} - \bs{\Omega}_{B_{j+1}}^I\bs{s}_{B_{j+1}J_j}
\end{align}
where the underlined terms are always zero. Rearranging gives the velocity constraint equation for joint $j$ as follows:
\begin{align}
  \bs{v}_{B_{j+1}}^I - \bs{v}_{B_j}^I - \mat{S}_{B_j J_j}\bs{\omega}_{B_j}^I + \mat{S}_{B_{j+1} J_j}\bs{\omega}_{B_{j+1}}^I = \mat{0}  .
\end{align}
Taking the inertial time derivative of the velocity constraint yields the acceleration constraint:
\begin{align}
    D^I\bs{v}_{B_{j+1}}^I - D^I\bs{v}_{B_j}^I - \mat{S}_{B_j J_j}D^I\bs{\omega}_{B_j}^I + \mat{S}_{B_{j+1} J_j}D^I\bs{\omega}_{B_{j+1}}^I = \bs{\gamma}_j .
\end{align}
The right-hand side, $\bs{\gamma}_j \in \mathbb{R}^3$, contains all terms dependent on velocity products $    \bs{\gamma}_j = \bs{\Omega}_{B_{j+1}}^I \bs{\Omega}_{B_{j+1}}^I \bs{s}_{B_{j+1} J_j}-\bs{\Omega}_{B_j}^I \bs{\Omega}_{B_j}^I \bs{s}_{B_j J_j} 
$. These $N-1$ first-order tensor constraints can be expressed in the second-order form as:
\begin{align}
    \mat{J} D^I\bs{\nu} = \bs{\gamma}
\end{align}
where $\bs{\nu} = [\bs{\nu}_v^\intercal, \bs{\nu}_\omega^\intercal]^\intercal \in \mathbb{R}^{6N}$ is the generalized velocity twist,  $\bs{\gamma} = [\bs{\gamma}_1^\intercal, \ldots, \bs{\gamma}_{N-1}^\intercal]^\intercal \in \mathbb{R}^{3(N-1)}$ is the stack of velocity-product terms, and $\mat{J} = [\mat{J}_v, \mat{J}_\omega] \in \mathbb{R}^{3(N-1) \times 6N}$ is the system's constraint Jacobian matrix. The block components of the Jacobian are $\mat{J}_v \in \mathbb{R}^{3(N-1) \times 3N}$ and $\mat{J}_\omega \in \mathbb{R}^{3(N-1) \times 3N}$ with the following structures:
\begin{align}
   \mat{J}_v = \begin{bmatrix}
-\mat{I}_3 & \mat{I}_3 & \mat{0} & \cdots & \mat{0} \\
\mat{0} & -\mat{I}_3 & \mat{I}_3 & & \vdots \\
\vdots & & \ddots & \ddots & \mat{0} \\
\mat{0} & \cdots & \mat{0} & -\mat{I}_3 & \mat{I}_3
\end{bmatrix} 
\end{align}
\begin{align}
    \mat{J}_\omega = \begin{bmatrix}
-\mat{S}_{B_1 J_1} & \mat{S}_{B_2 J_1} & \mat{0} & \cdots & \mat{0} \\
\mat{0} & -\mat{S}_{B_2 J_2} & \mat{S}_{B_3 J_2} & & \vdots \\
\vdots & & \ddots & \ddots & \mat{0} \\
\mat{0} & \cdots & \mat{0} & -\mat{S}_{B_{N-1} J_{N-1}} & \mat{S}_{B_N J_{N-1}}
\end{bmatrix} .
\end{align}

\subsection{Assembled System Equations}
The dynamics and kinematics can be combined into a single augmented system of Differential-Algebraic Equations (DAEs). The internal joint forces $\mat{F}_J$ act as Lagrange multipliers that enforce the kinematic constraints. Therefore, the unreduced equations of motion for the constrained multibody system in closed form can be written as follows:
\begin{align}\label{equ:tensor_mat_dyn}
 \begin{bmatrix}
\mathcal{M} & \mat{J}^\intercal \\
\mat{J} & \mat{0}
\end{bmatrix}
\begin{bmatrix}
D^I\bs{\nu} \\
-\mat{F}_J
\end{bmatrix}
=
\begin{bmatrix}
\mathcal{F} \\
\bs{\gamma}
\end{bmatrix}   .
\end{align}
Note the relationship $\mat{C}_F = -\mat{J}_v^\intercal$ and $\mat{C}_M = -\mat{J}_\omega^\intercal$. The term $-\mat{J}^\intercal \mat{F}_J$ represents the generalized constraint forces applied to the bodies. The components of this augmented system are the generalized mass matrix $\mathcal{M} = \operatorname{blockdiag}(\mat{M}_v, \mat{M}_\omega) \in \mathbb{R}^{6N \times 6N}$, and the generalized applied force tensor $\mathcal{F} \in \mathbb{R}^{6N}$, which includes all external interactions such as control force and moments, gravitational, gyroscopic, and internal joint moments (but not the constraint forces $\mat{F}_J$):
\begin{align}\label{equ:tensor_mat_map}
\mathcal{F} =
\begin{bmatrix}
\mat{F}_B + \mat{M}_v \mat{g}_{3N} \\
\mat{M}_B + \mat{C}_F \mat{M}_J - \bs{\mathcal{C}}(\bs{\nu})\bs{\nu}
\end{bmatrix}
\end{align}

The geometric tensors that form the Jacobian provide a physically interpretable structure for the system's internal dynamics and external force distributions. This clarity is advantageous for both the LHS control and RHS learning components, which reduces the complexity of the learning task and improves the system's adaptability. This linear system can be solved at each time step for the accelerations $D^I\bs{\nu}$ and the internal forces $\mat{F}_J$, allowing for the numerical integration of the system's state over time. 

While the tensor-based formulation provides a compact and coordinate-free representation of the system dynamics, numerical simulation requires the assignment of a common reference frame. This process, referred to as coordination, involves expressing each tensor as a matrix of its components and introducing the necessary transformation matrices to ensure all vectors are consistent. The attitude of each body, which determines these transformations, is tracked using a robust quaternion-based approach to avoid the singularities associated with Euler angles. The details of the quaternion kinematic equations and their integration are provided in Appendix~\ref{sec:app-att_det}. It is important to note that while the underlying tensor equations remain invariant, their coordinated form is symbolically more complex and significantly less readable. To illustrate this effect, the complete set of coordinated dynamic equations for a 3-body system is derived in Appendix~\ref{app:n3_example}. 
\begin{remark}
    In the formulation of constrained multibody dynamics, the equations of motion often take the form of a linear system involving an augmented or Karush-Kuhn-Tucker (KKT) matrix: 
\begin{align}
 \mat{K} = \begin{bmatrix}
\mathcal{M} & \mat{J}^\intercal \\
\mat{J} & \mat{0}
\end{bmatrix}   .
\end{align}
The matrix $\mat{K}$ is invertible if and only if the following two conditions are met. For any physically meaningful system, these conditions are generally satisfied.
\begin{itemize}
    \item  The mass matrix $\mathcal{M}$ must be symmetric and positive definite. This is always true for systems composed of bodies with positive mass, as kinetic energy must be positive for any non-zero velocity. The symmetric and positive definiteness is mathematically evident.
    \item  The constraint Jacobian $\mat{J}$ must have full row rank. This implies that all defined kinematic constraints are independent, meaning none are redundant or contradictory. In current multibody, mathematically it is straightforward to show that for any vector $\bs{c} \in \mathbb{R}^{3(N-1)}$, the equation $\bs{c}^\intercal \mat{J} = \bs{0}$ implies that $\bs{c} = \bs{0}$, exploiting only linear Jacobian $\bs{J}_v$.
\end{itemize}
Failure to meet these conditions, particularly the second, points to an ill-posed physical model, which would not have a unique solution for its accelerations and constraint forces. When the aforementioned conditions are met, the inverse of the augmented matrix $\mat{K}$ can be expressed analytically. The inverse is given by:
\begin{align}
  \mat{K}^{-1} = 
\begin{bmatrix}
\mathcal{M} & \mat{J}^\intercal \\
\mat{J} & \mat{0}
\end{bmatrix}^{-1} = 
\begin{bmatrix}
\mathcal{M}^{-1} - \mathcal{M}^{-1}\mat{J}^\intercal \mat{S}^{-1} \mat{J}\mathcal{M}^{-1} & \mathcal{M}^{-1}\mat{J}^\intercal \mat{S}^{-1} \\
\mat{S}^{-1}\mat{J}\mathcal{M}^{-1} & -\mat{S}^{-1}
\end{bmatrix}  
\end{align}
The formula depends on the matrix $\mat{S}$, which is the Schur complement of $\mathcal{M}$: $\mat{S} = \mat{J}\mathcal{M}^{-1}\mat{J}^\intercal$.  $\mat{S}$ is often interpreted as the operational space inertia matrix, representing the effective inertia of the system at the points of constraint. The conditions for the invertibility of $\mat{K}$ guarantee that $\mat{S}$ is also invertible.
\end{remark}

\section{Tensor Invariant Data-Assisted Control}\label{sec:control}
The tensor-based equations of motion provide the ideal foundation for the DAC framework. First, the system is decomposed into a well-modeled LHS and an uncertain RHS, connected by a virtual port variable, $\bs{\tau}$, which represents the total generalized forces and moments acting on the system's mechanical structure. The augmented dynamics are rearranged to isolate this port:
\begin{equation}\label{equ:tensor_mat_dyn_decomposed}
\underbrace{
    \begin{bmatrix}
        \mathcal{M} & \mat{J}^\intercal \\
        \mat{J} & \mat{0}
    \end{bmatrix}
    \begin{bmatrix}
        D^I\bs{\nu} \\
        -\mat{F}_J
    \end{bmatrix}
    -
    \begin{bmatrix}
        \bs{\mathcal{F}}_{known} \\
        \bs{\gamma}
    \end{bmatrix}
}_{\text{LHS: Well-Modeled Dynamics}}
=
\underbrace{
    \begin{bmatrix}
        \bs{\tau} \\
        \mat{0}
    \end{bmatrix}
}_{\text{Interaction Port}} = \underbrace{
    \begin{bmatrix}
        \begin{bmatrix}
        \bs{F}_B \\
        \bs{M}_B\\
    \end{bmatrix}\\
    \mat{0}\\
    \end{bmatrix}
}_{\text{RHS: Uncertain Model (Map)}}
\end{equation}
where $\bs{\mathcal{F}}_{known} = \mat{M}_v \bs{g}_{3N} - \bs{\mathcal{C}}(\bs{\nu})\bs{\nu} + \mat{C}_F \mat{M}_J$ contains all well-known and structurally exact terms. The crucial insight is that the RHS represents external control, dissipation, and disturbances, corresponding to the physical inputs $\mat{F}_B$ and $\mat{M}_B$ (in general, any interaction with the environment). The control strategy is to design a model-based law for a desired port value, $\bs{\tau}_c$, to stabilize the LHS and track desired trajectories, which then becomes the setpoint for the RHS controller. 

This work has presented three hierarchical control strategies: \textbf{(I)} velocity control of all bodies, \textbf{(II)} combined position and velocity control of all bodies, and \textbf{(III)} task-space control of the end-effector (Body N) with null-space obstacle avoidance. The focus has been on rigorously designing the model-based LHS controller to guarantee stability and performance for these objectives. The design of the data-assisted RHS controller, which would learn the uncertain mapping from the desired port variable to the physical actuator inputs, remains a topic for future work. This paper serves as the foundational study, establishing the necessary coordinate-free control laws.

The use of unreduced, tensor-based dynamics provides a compelling foundation for future learning-based control. This approach creates a clean and principled separation between the structurally known LHS and the uncertain RHS, which is simplified to an algebraic mapping. This simplified structure makes the RHS an ideal target for advanced, data-efficient machine learning techniques, such as equivariant learning. The central element bridging these two domains is the coordinate-free control law designed herein. It provides a robust, theoretically guaranteed command signal that serves as a consistent and abstract target for any subsequent RHS learning algorithm, ensuring a stable and modular DAC architecture.

\subsection{(I) Velocity Control of Bodies}
The objective is to ensure the system's generalized velocity tensor, $\bs{\nu}$, accurately tracks a desired trajectory, $\bs{\nu}_d(t)$. To do this, a surface error, $\bs{s}$, representing the velocity tracking error $\bs{s} = \bs{\nu} - \bs{\nu}_d$ is defined. The desired trajectory $\bs{\nu}_d$ must be physically feasible, meaning it must satisfy the kinematic constraints of the system, i.e., $\mat{J}\bs{\nu}_d = \bs{0}$ (the constraint Jacobian on velocity and acceleration are the same). Consequently, the error tensor $\bs{s}$ is also constrained and must lie in the null space of the Jacobian: $\mat{J}\bs{s} = \mat{J}(\bs{\nu} - \bs{\nu}_d) = \bs{0}$. 
This property is fundamental to the control design, as it ensures the error dynamics evolve on the valid motion manifold of the system.

The control law for the desired port variable $\bs{\tau}_c$ to drive the sliding error $\bs{s}$ to zero can be defined only for the $6N$ generalized forces and moments that comprise the port:
\begin{equation}\label{equ:smc_law}
    \bs{\tau}_c = \mathcal{M} (D^I \bs{\nu}_d) + \bs{\mathcal{C}}(\bs{\nu})\bs{\nu}_d - \bs{\mathcal{F}}_{g,J} - \mat{K}_d \bs{s}
\end{equation}

where $\mat{K}_d$ is a symmetric, positive-definite gain second-order tensor, and $\bs{\mathcal{F}}_{g,J} = \mat{M}_v \bs{g}_{3N}  + \mat{C}_F \mat{M}_J$. This law computes the necessary generalized force to counteract the known system dynamics, achieve the desired acceleration, and add a corrective term that is proportional to the error.

To prove convergence in a coordinate-free manner,  a Lyapunov candidate function based on the kinetic energy of the error state is defined:
\begin{equation}
    V = \frac{1}{2}\bs{s}^\intercal \mathcal{M} \bs{s}.
\end{equation}
The rotational time derivative of this scalar function is:
\begin{equation}
    D^I V = \bs{s}^\intercal \mathcal{M} (D^I\bs{s}) + \frac{1}{2}\bs{s}^\intercal (D^I\mathcal{M})\bs{s}.
\end{equation}

For the design of the LHS controller, we assume the RHS can perfectly realize the commanded port value, such that $\bs{\tau} = \bs{\tau}_c$. Substituting the control law \eqref{equ:smc_law}, yields the following closed-loop error dynamics:
\begin{equation}\label{equ:cl_s}
    \mathcal{M}(D^I\bs{s}) + \bs{\mathcal{C}}(\bs{\nu})\bs{s} = -\mat{K}_d\bs{s} + \mat{J}^\intercal\mat{F}_J . 
\end{equation}

Hence, 
\begin{equation}
    D^I V = \bs{s}^\intercal \left(-\mat{K}_d\bs{s} + \mat{J}^\intercal\mat{F}_J - \bs{\mathcal{C}}(\bs{\nu})\bs{s}\right)  + \frac{1}{2}\bs{s}^\intercal (D^I\mathcal{M})\bs{s} .
\end{equation}
Since $\mat{J}\bs{s}=\bs{0}$, the term involving the unknown internal force $\mat{F}_J$ vanishes:
$
\bs{s}^\intercal \mat{J}^\intercal\mat{F}_J = 0
$. Also always $\bs{s}^\intercal\left( D^I\mathcal{M}-2\bs{\mathcal{C}}\right)\bs{s}=0$, see Appendix \ref{app:m_dot_2c}. This leaves the final, coordinate-free result $
    D^I V = -\bs{s}^\intercal \mat{K}_d \bs{s}
$. When this scalar result is expressed in any fixed coordinate system, such as the inertial frame $I$, the inertial derivative of the scalar Lyapunov function becomes the simple time derivative, i.e., $[D^I V]^I = \dot{V}$. The $\mat{K}_d$ can be selected as a positive-definite gain matrix in the inertial frame, and hence $\dot{V} \le 0$, which proves that the sliding error $\bs{s}$ asymptotically converges to zero.

\subsection{(II) Control Law for Position and Attitude Tracking}
The previous control law guarantees velocity tracking. To ensure the system converges to a desired state in both position and attitude, a new composite error in the sliding surface should be introduced. The desired state is given by the body positions $\bs{s}_{B_iI,d}$, attitudes as quaternions $\bs{q}_d^{B_iI}$, linear velocities $\bs{v}_{B_i,d}^I$, and angular velocities $\bs{\omega}_{B_i,d}^I$. Then, the generalized position error,  $\bs{e}_p$, for all $N$ bodies is defined as:
\begin{align}
 \bs{e}_p = \begin{bmatrix} \bs{e}_{pos} \\ \bs{e}_{att} \end{bmatrix}   
\end{align}
where the linear position error is the straightforward difference:
\begin{align}
\bs{e}_{pos} = \begin{bmatrix} \bs{s}_{B_1I} - \bs{s}_{B_1I,d} \\ \vdots \\ \bs{s}_{B_NI} - \bs{s}_{B_NI,d} \end{bmatrix}      
\end{align}
and the attitude error is best described using quaternions to avoid singularities. The error quaternion $\bs{q}_e$ representing the rotation from the current attitude $\bs{q}$ to the desired attitude $\bs{q}_d$ is found by quaternion multiplication: $\bs{q}_e = (\bs{q}_d)^{-1} \otimes \bs{q}$. The vector part of this error quaternion, $\bs{q}_{e,v}$, serves as an error tensor:
\begin{align}
         \bs{e}_{att} = \begin{bmatrix} \bs{q}_{e,v}^{B_1I} \\ \vdots \\ \bs{q}_{e,v}^{B_NI} \end{bmatrix} . 
\end{align}
Generalized velocity error, $\bs{e}_v$, is the sliding surface used previously, i.e. 
$\bs{e}_v = \bs{\nu} - \bs{\nu}_d$.  Now, a new sliding surface $\bs{s}$ that is a weighted sum of the velocity error and the position/attitude error should be defined:
\begin{equation}
    \bs{s} = \bs{e}_v + \mat{\Lambda} \bs{e}_p = (\bs{\nu} - \bs{\nu}_d) + \mat{\Lambda}\bs{e}_p
\end{equation}
where $\mat{\Lambda}$ is a symmetric, positive-definite gain matrix, typically block-diagonal: $\mat{\Lambda} = \operatorname{blockdiag}(\mat{\Lambda}_p, \mat{\Lambda}_a)$. This can be rewritten as $\bs{s} = \bs{\nu} - \bs{\nu}_r$, where a reference velocity $\bs{\nu}_r$ is defined: $\bs{\nu}_r = \bs{\nu}_d - \mat{\Lambda}\bs{e}_p$.  The reference velocity $\bs{\nu}_r$ is the velocity the system should have at any instant to ensure the position error $\bs{e}_p$ decays exponentially. The control objective is now to make the true velocity $\bs{\nu}$ track this reference velocity $\bs{\nu}_r$.

The control law for the desired port variable $\bs{\tau}_c$ is designed to drive the new composite sliding surface $\bs{s} = \bs{\nu} - \bs{\nu}_r$ to zero. The law has a similar structure to before, but targets the reference velocity's acceleration:
\begin{align}\label{equ:smc_law_pos}
    \bs{\tau}_c = \mathcal{M} (D^I \bs{\nu}_r) + \bs{\mathcal{C}}(\bs{\nu})\bs{\nu}_r  - \bs{\mathcal{F}}_{g,J} - \mat{K}_d \bs{s} .
\end{align}
The required inertial derivative of the reference velocity, $D^I \bs{\nu}_r$, is computable from the system state:
\begin{align}
 D^I \bs{\nu}_r = D^I\bs{\nu}_d - \mat{\Lambda}(D^I\bs{e}_p)  . 
\end{align}

The derivative of the position error $D^I\bs{e}_p$ is simply the velocity error, $\bs{e}_v = \bs{\nu} - \bs{\nu}_d$. The derivative of the attitude error is a known kinematic function of the angular velocity error. Therefore, $D^I \bs{\nu}_r$ is a known function of the current state and desired trajectory.

The stability proof follows the same path as the velocity controller. Using the same Lyapunov candidate function of new $\bs{s}$:
\begin{equation}
    V = \frac{1}{2}\bs{s}^\intercal \mathcal{M} \bs{s} .
\end{equation}
The derivation for its rotational time derivative, $D^I V$, is identical to the previous. Again, it is assumed the RHS can perfectly realize the commanded port value, such that $\bs{\tau} = \bs{\tau}_c$. Substituting the new control law \eqref{equ:smc_law_pos}, yields again the same closed-loop error dynamics \eqref{equ:cl_s}. The control law is structured to cancel the complex dynamic terms, and the term with the unknown constraint forces $\mat{F}_J$ vanishes because the sliding surface $\bs{s}$ lies on the constraint-valid motion manifold (i.e., $\mat{J}\bs{s} = \bs{0}$). This leads to the same, coordinate-free result: $D^I V = -\bs{s}^\intercal \mat{K}_d \bs{s}$. 
When expressed in the inertial frame, this becomes $\dot{V} = -\bs{s}^\intercal \mat{K}_d \bs{s} \le 0$, and hence $\bs{s} \to \bs{0}$ asymptotically.  Since $\bs{s} = (\bs{\nu} - \bs{\nu}_d) + \mat{\Lambda}\bs{e}_p$, having $\bs{s} \to \bs{0}$ means:
\begin{align}
   (\bs{\nu} - \bs{\nu}_d) + \mat{\Lambda}\bs{e}_p = \bs{0}. 
\end{align}
Recognizing that $\bs{\nu} - \bs{\nu}_d = D^I \bs{e}_p$, it is left with the error dynamics:
\begin{align}
   D^I\bs{e}_p + \mat{\Lambda}\bs{e}_p = \bs{0} . 
\end{align}
This is a stable, first-order linear differential equation for the position and attitude error $\bs{e}_p$. Since $\mat{\Lambda}$ is positive-definite, the solution to this equation is an exponential decay to zero. Therefore, this control law not only guarantees that the velocity error converges to zero, but also that the linear position and attitude errors converge to zero exponentially.

\subsection{(III) Control Law for End-Effector Tracking with Null-Space Obstacle Avoidance}
The previous control law was designed to control the full state of all $N$ bodies. A more practical objective is controlling only the position and attitude of the end-effector (attached to body $N$), while using the remaining degrees of freedom (the null space of the primary task) for obstacle avoidance.

Task space is defined as the 6D space of the end-effector's motion and orientation. The end-effector's velocity, $\bs{\nu}_{e} = [(\bs{v}_{B_N}^I)^\intercal, (\bs{\omega}_{B_N}^I)^\intercal]^\intercal$, is related to the full system's generalized velocity, $\bs{\nu}$, by a task Jacobian, $\mat{J}_{t}$ as: $
    \bs{\nu}_{e} = \mat{J}_{t} \bs{\nu}
$. For this specific task, $\mat{J}_{t}$ is a simple $6 \times 6N$ selection matrix:
\begin{align}
\mat{J}_{t} = \begin{bmatrix} \mat{0} & \cdots & \mat{0} & \mat{I}_3 & \mat{0} & \cdots & \mat{0} \\ \mat{0} & \cdots & \mat{0} & \mat{0} & \mat{0} & \cdots & \mat{I}_3 \end{bmatrix}    
\end{align}
where the identity matrices select the linear and angular velocities of body $N$. The control objective is now defined entirely in the task space. The end-effector position/attitude error ($\bs{e}_{e,p}$)  is a $6 \times 1$ tensor containing the linear position error $(\bs{s}_{B_NI} - \bs{s}_{B_NI,d})$ and the attitude error ($\bs{q}_{e,v}^{B_NI}$) of the end-effector. And, the end-effector velocity error ($\bs{e}_{e,v}$) is defined as $\bs{e}_{e,v} = \bs{\nu}_{e} - \bs{\nu}_{e,d}$. A task-space sliding surface, $\bs{s}_{t}$, as a composite error is defined:
\begin{align}
    \bs{s}_{t} = \bs{e}_{e,v} + \mat{\Lambda}_{e} \bs{e}_{e,p}
\end{align}
where $\mat{\Lambda}_{e}$ is a $6 \times 6$ positive-definite gain matrix. This can be rewritten as $\bs{s}_{t} = \bs{\nu}_{e} - \bs{\nu}_{e,r}$, where the task-space reference velocity is: $
 \bs{\nu}_{e,r} = \bs{\nu}_{e,d} - \mat{\Lambda}_{e}\bs{e}_{e,p}   
$.

The system has more degrees of freedom than the task requires, creating a null space. We can command motion in this null space without affecting the end-effector's trajectory \cite{khatib2003unified}. The pseudo-inverse of the task Jacobian, weighted by the system's mass tensor, can be written as:
\begin{align}
   \mat{J}_{t}^\dagger = \mathcal{M}^{-1}\mat{J}_{t}^\intercal(\mat{J}_{t}\mathcal{M}^{-1}\mat{J}_{t}^\intercal)^{-1}. 
\end{align}
Then, the mapping that projects an arbitrary velocity into the null space will be $
    \mat{P} = \mat{I} - \mat{J}_{t}^\dagger \mat{J}_{t}$. Now, the potential field $U_{o}(\bs{s}_{B_1I}, \dots, \bs{s}_{B_{N-1}I})$, is formulated so that the desired velocity move away from obstacles, in the negative direction of gradient of this potential, projecting the velocities into the null space 
    $
    \bs{\nu}_{n} = -\mat{P} \bs{\nabla} U_{o} .
    $

The primary task and the secondary objective to define a single reference velocity, $\bs{\nu}_r$, for the entire system is combined as follows:
\begin{equation}
    \bs{\nu}_r = \mat{J}_{t}^\dagger \bs{\nu}_{e,r} + \bs{\nu}_{n}.
\end{equation}
This composite reference velocity simultaneously commands the end-effector to follow its trajectory and the rest of the system to avoid obstacles. The final control law for the port variable $\bs{\tau}_c$ is designed to make the full system velocity $\bs{\nu}$ track this new composite reference velocity $\bs{\nu}_r$. The sliding surface is the same as equations in (II) and $\bs{s} = \bs{\nu} - \bs{\nu}_r$, with the same control law \eqref{equ:smc_law_pos}. The Lyapunov stability proof is identical to the previous case. It guarantees the end-effector's velocity converges to its reference velocity, $\bs{\nu}_{e} \to \bs{\nu}_{e,r}$. This leads to the task-space error dynamics $D^I\bs{e}_{e,p} + \mat{\Lambda}_{e}\bs{e}_{e,p} = \bs{0}$, ensuring the end-effector's position and attitude converge to their desired values. Also, it ensures the component of the system's motion in the null space converges to the desired obstacle avoidance velocity, $\mat{P}\bs{\nu} \to \bs{\nu}_{n}$.

\section{Simulation and Validation}\label{sec:simulation}
This section validates the derived tensor-based model and the control law through numerical simulations. It focuses on the most comprehensive controller developed, Control Law (III). The validation is performed in two sequential stages. First, the open-loop dynamics are verified to ensure the physical and numerical correctness of the model. Once the open-loop model is validated, the performance of the closed-loop controller is evaluated on a more complex scenario.

\subsection{Open-Loop Validation}
The open-loop tests are conducted on a 5-body chain to clearly demonstrate the fundamental dynamic behaviors. A specific sequence of external forces and moments is applied via the port variable $\bs{\tau}$ to excite the system, and the simulation data is analyzed against the following criteria:

\begin{enumerate}
    \item \textbf{Conservation of Momentum:} During periods where no external forces or moments are applied ($\bs{\tau} = \bs{0}$), the total linear and angular momentum of the system must be conserved. This provides a powerful check on the correctness of the overall model and, in particular, the mass matrix, inertia transformations, and gyroscopic terms.
    \item \textbf{Holonomic Constraint Integrity:} The kinematic constraints at the joints must be satisfied. To verify it, the constraint violation error, i.e., the positional mismatch at each joint representing the integral of the velocity constraint, is monitored. A significant growth in this error indicates an error in the model, and small fraction drifts are due to numerical errors.    
    \item \textbf{Internal Force Behavior:} The behavior of the internal joint forces, $\mat{F}_J$, provides a physical sanity check. When the system is subjected to a pure external moment, $\mat{F}_J$ should remain near zero. Conversely, when a translational external force is applied, non-zero, continuous, and bounded internal forces must be generated to propagate this force through the chain.
\end{enumerate}

\begin{figure}[ht]
    \centering
    \includegraphics[width=0.7\linewidth]{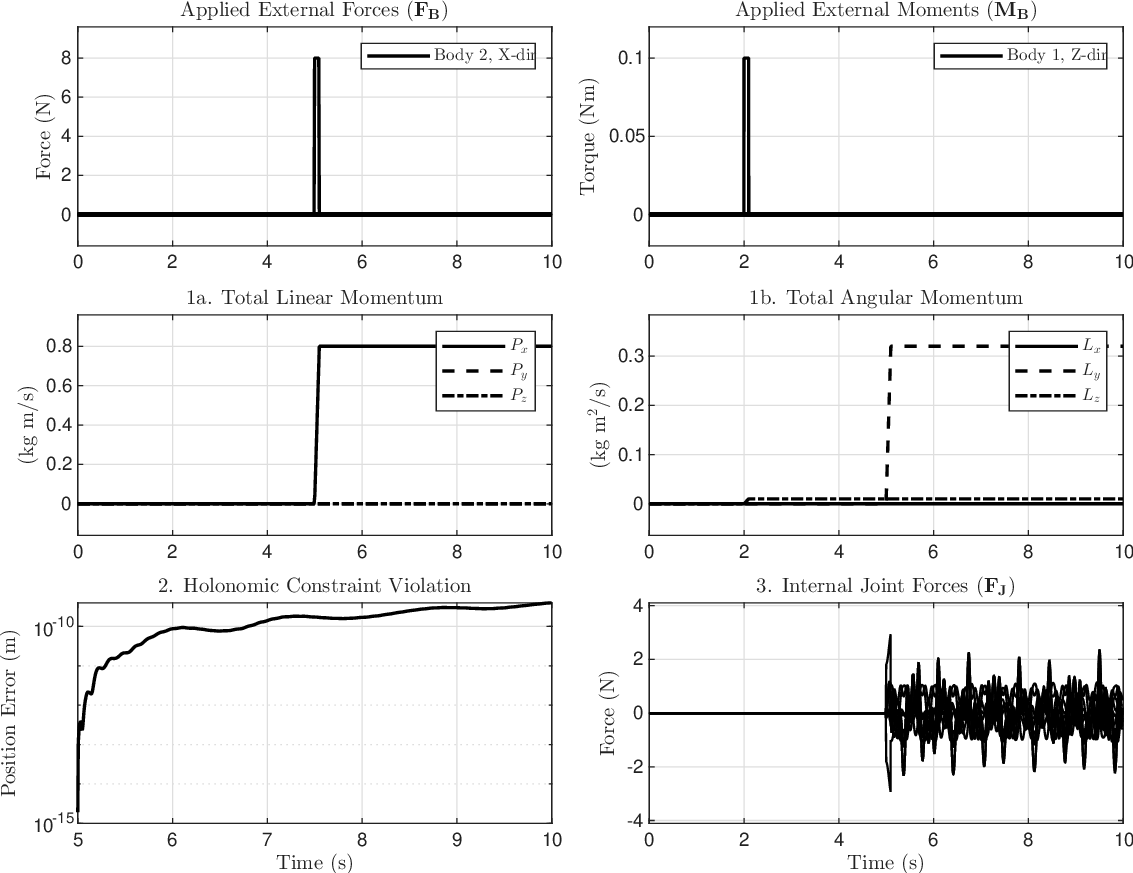}
    \caption{Validation of open loop system dynamics}
    \label{fig:validation_figure}
\end{figure}

\begin{figure}[ht]
    \centering
    \includegraphics[width=0.7\linewidth]{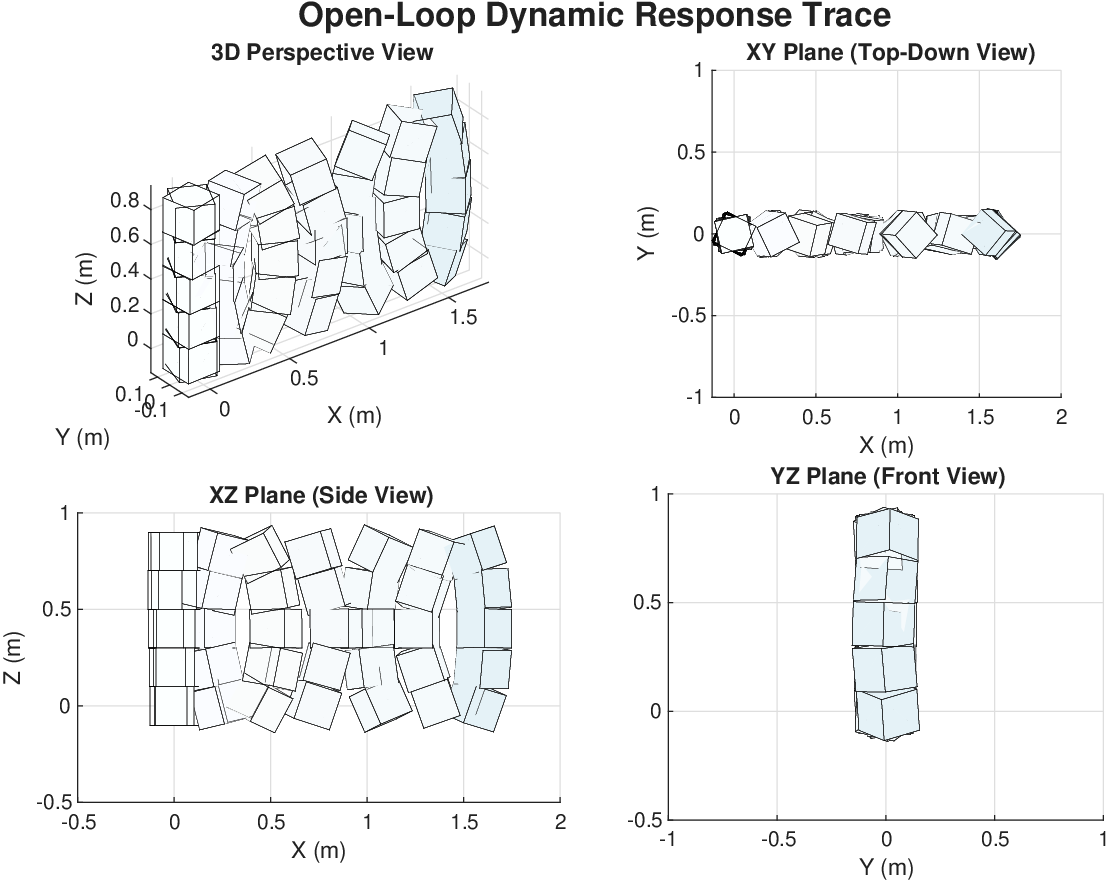}
    \caption{A trace of the multibody movement under the influence of the applied external force and moment pulses.}
    \label{fig:trace_figure}
\end{figure}

Figure~\ref{fig:validation_figure} presents the results of the open-loop validation test for the 5-body system, confirming that the model successfully passes all three validation criteria. Figure~\ref{fig:trace_figure} shows the traced animation of the multibody motions graphically. For the sake of demonstrating all bodies in a more compact view, gravity forces are compensated to restrict large movement in the Z-direction.

The simulation correctly demonstrates the conservation of momentum. Initially at rest, both total linear and angular momentum are zero. When the moment pulse is applied to Body 1, the total angular momentum increases accordingly and then remains constant after the moment is removed. Likewise, the subsequent force pulse on Body 2 increases the total linear momentum, which is also conserved after the force vanishes. Notably, the model correctly captures the physical coupling where the translational force induces a change in the total angular momentum, as the force creates a torque about the system's overall center of mass.  The holonomic constraint violation remains negligible throughout the simulation, confirming numerical stability. The internal joint forces ($\mat{F}_J$) exhibit plausible behavior: they remain zero during the pure moment pulse and become non-zero with stable, bounded values to correctly propagate the external force through the chain.

\subsection{Closed-Loop Validation}
\begin{figure}[ht]
    \centering
    \includegraphics[width=0.7\linewidth]{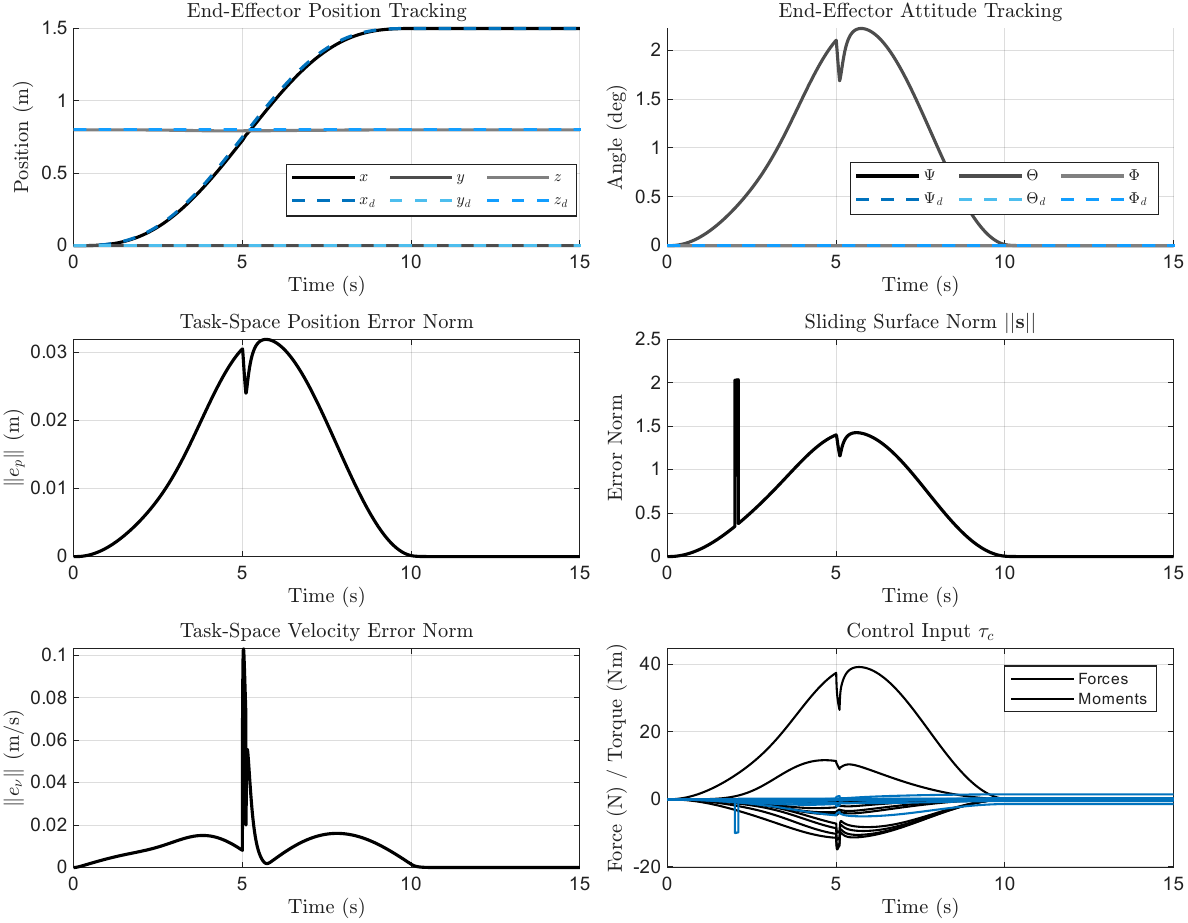}
    \caption{Closed-loop performance of the end-effector tracking controller, showing desired (dashed-line) vs. actual trajectories and error norms.}
    \label{fig:closed_loop_performance}
\end{figure}

\begin{figure}[ht]
    \centering
    \includegraphics[width=0.7\linewidth]{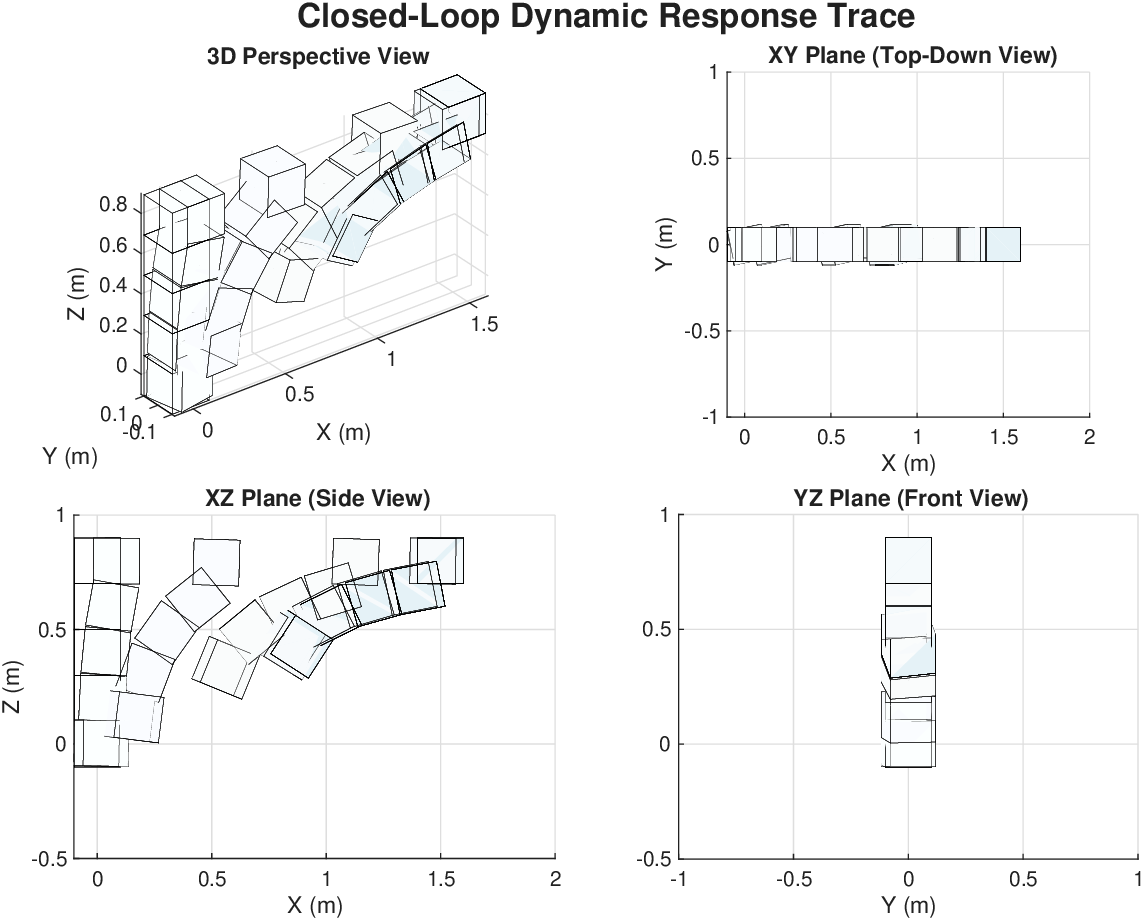}
    \caption{Trace of multibody movement in closed-loop}
    \label{fig:trace_closed_loop}
\end{figure}

After verifying the open-loop model, the controller is enabled to evaluate its performance. Figure~\ref{fig:closed_loop_performance} shows the simulation results, demonstrating that the end-effector successfully tracks its desired trajectory. External disturbances are introduced to verify that the controller rejects them while maintaining stability. Figure~\ref{fig:trace_closed_loop} shows the trace of the system's motion during the closed-loop simulation.

\section{Conclusion}\label{sec:conclusion}

This paper presented a comprehensive framework for the modeling and control of complex, constrained multibody systems, founded on the principles of tensor mechanics. By employing a non-recursive Newton-Euler formulation, the complete system dynamics in an abstract, tensor invariant form are derived. This rigorous model provides an ideal structure for the Data-Assisted Control framework, enabling a clear and physically meaningful separation between the well-modeled, conservative dynamics (LHS) and the uncertain interaction port (RHS). Then, a series of tensor-based control laws for LHS are developed, culminating in a sophisticated task-space law for end-effector tracking. Formal Lyapunov analysis, performed in a coordinate-free manner, rigorously proved the stability and convergence of these controllers, guaranteeing performance based on the physical properties of the system.

The theoretical framework was validated through simulation. Open-loop tests confirmed the model's physical and numerical integrity by verifying the conservation of momentum and the satisfaction of holonomic constraints under external inputs. Subsequent closed-loop simulations demonstrated the controller's effectiveness in achieving high-performance trajectory tracking for the end-effector. By designing the control law at the tensor level, a robust and abstract foundation for the control of multibody systems is established. The resulting commanded port variable, $\bs{\tau}_c$, serves as a consistent and theoretically-guaranteed target signal. This provides a stable and well-defined interface for future work in designing the data-driven RHS controller, making it an ideal platform for advanced machine learning techniques to handle real-world uncertainties in dissipative terms, control inputs, and unmodeled interaction forces.

\section*{Declarations}

 \begin{itemize}
\item This work was not supported by any organization.

\end{itemize}

\appendix
\section{Attitude Determination}\label{sec:app-att_det}
To make the tensor-based equations numerically solvable, the attitude of each body must be tracked to compute the necessary transformation matrices. To avoid the singularities associated with Euler angles, quaternions are used to represent and propagate the attitude of each body.

\subsection{Quaternion Kinematics}
For each body $i$, we define a quaternion $\bs{q}^{B_iI} = [q_0, q_1, q_2, q_3]^\intercal$ that represents the orientation of the body frame $B_i$ with respect to the inertial frame $I$. The time evolution of this quaternion is governed by the body's angular velocity. Let the angular velocity of body $i$ with respect to the inertial frame, expressed in its own body coordinates, be $\bs{\omega}_{B_i}^{I} = [p^{B_iI}, q^{B_iI}, r^{B_iI}]^\intercal$. The quaternion differential equation is then:
\begin{equation}\label{eq:diff_q}
\begin{Bmatrix} \dot{q}_0^{B_iI} \\ \dot{q}_1^{B_iI} \\ \dot{q}_2^{B_iI} \\ \dot{q}_3^{B_iI} \end{Bmatrix}
= \frac{1}{2}
\begin{bmatrix}
 0 & -p^{B_iI} & -q^{B_iI} & -r^{B_iI} \\
 p^{B_iI} & 0 & r^{B_iI} & -q^{B_iI} \\
 q^{B_iI} & -r^{B_iI} & 0 & p^{B_iI} \\
 r^{B_iI} & q^{B_iI} & -p^{B_iI} & 0
\end{bmatrix}
\begin{Bmatrix} {q}_0^{B_iI} \\ {q}_1^{B_iI} \\ {q}_2^{B_iI} \\ {q}_3^{B_iI} \end{Bmatrix}
\end{equation}
This system of four linear, first-order ordinary differential equations is integrated numerically for each body to track its attitude over time.

\subsection{Initialization and Conversion}
To start the integration, the initial quaternion for each body is calculated from its initial set of Euler angles $(\Phi_i, \Theta_i, \Psi_i)$, which represent a Z-Y-X (Yaw, Pitch, Roll) rotation sequence.
\begin{align}\label{eq:init_q}
  q_0^{B_iI}(0) &= \cos(\Psi_i/2)\cos(\Theta_i/2)\cos(\Phi_i/2)+\sin(\Psi_i/2)\sin(\Theta_i/2)\sin(\Phi_i/2)\nonumber\\
  q_1^{B_iI}(0) &= \cos(\Psi_i/2)\cos(\Theta_i/2)\sin(\Phi_i/2)-\sin(\Psi_i/2)\sin(\Theta_i/2)\cos(\Phi_i/2)\nonumber\\
  q_2^{B_iI}(0) &= \cos(\Psi_i/2)\sin(\Theta_i/2)\cos(\Phi_i/2)+\sin(\Psi_i/2)\cos(\Theta_i/2)\sin(\Phi_i/2)\nonumber\\
  q_3^{B_iI}(0) &= \sin(\Psi_i/2)\cos(\Theta_i/2)\cos(\Phi_i/2)-\cos(\Psi_i/2)\sin(\Theta_i/2)\sin(\Phi_i/2)
\end{align}
Once the quaternion $\bs{q}^{B_iI}$ is known at any time, it can be converted back to Euler angles or used to construct the transformation matrix $[\mat{T}]^{B_iI}$, which transforms a tensor from inertial coordinates to body coordinates.
\begin{align}\label{eq:euler_q}
\Phi_i &= \operatorname{atan2} \left(2(q_2^{B_iI}q_3^{B_iI}+q_0^{B_iI}q_1^{B_iI}), (q_0^{B_iI})^2-(q_1^{B_iI})^2-(q_2^{B_iI})^2+(q_3^{B_iI})^2\right)\nonumber\\
\Theta_i &= \arcsin(-2(q_1^{B_iI}q_3^{B_iI}-q_0^{B_iI}q_2^{B_iI}))\nonumber\\
\Psi_i &= \operatorname{atan2} \left(2(q_1^{B_iI}q_2^{B_iI}+q_0^{B_iI}q_3^{B_iI}), (q_0^{B_iI})^2+(q_1^{B_iI})^2-(q_2^{B_iI})^2-(q_3^{B_iI})^2\right)
\end{align}
\begin{equation}\label{eq:tm_q}
[\mat{T}]^{B_iI} =
\begin{bmatrix}
 (q_0^2+q_1^2-q_2^2-q_3^2) & 2(q_1q_2+q_0q_3) & 2(q_1q_3-q_0q_2) \\
 2(q_1q_2-q_0q_3) & (q_0^2-q_1^2+q_2^2-q_3^2) & 2(q_2q_3+q_0q_1) \\
 2(q_1q_3+q_0q_2) & 2(q_2q_3-q_0q_1) & (q_0^2-q_1^2-q_2^2+q_3^2)
\end{bmatrix}
\end{equation}
\textit{Note: Superscripts on quaternions in the matrix are omitted for brevity.}

\subsection{Relative Attitude}
The same procedure is used to determine the relative orientation between adjacent bodies, which is required for calculating the joint moments $\bs{m}_{J_j}$. To find the relative Euler angles $(\phi_j, \theta_j, \psi_j)$ between body $j$ and body $j+1$, one can solve the same quaternion kinematic equations, replacing the inertial frame $I$ with the reference frame $B_{j+1}$. The necessary input is the relative angular velocity, $\bs{\omega}^{B_j B_{j+1}}$, which is given by $\bs{\omega}^{B_j I} - \bs{\omega}^{B_{j+1} I}$ after ensuring both vectors are expressed in a common reference frame.

\section{Coordinate-Specific Augmented Matrix Formulation (N=3)} \label{app:n3_example}
This section provides the fully specified augmented matrix equations for the $N=3$ system. All vectors and matrices are explicitly assigned to a coordinate frame, all necessary transformation matrices are included, and inertial time derivatives are simplified to standard time derivatives.

\subsection{State and Force Vectors}
The unknown vectors are now expressed with their coordinate frames, and tensors are changed to vectors and matrices.
\begin{itemize}
    \item \textbf{Generalized Acceleration Vector} ($18 \times 1$): The time derivative of the state vector, with all components in the inertial frame $I$.
    \begin{align}
       \dot{\bs{\nu}}^I = 
    \begin{bmatrix}
    [\dot{\bs{v}}_{B_1}^I]^I \\ [\dot{\bs{v}}_{B_2}^I]^I \\ [\dot{\bs{v}}_{B_3}^I]^I \\ [\dot{\bs{\omega}}_{B_1}^I]^I \\ [\dot{\bs{\omega}}_{B_2}^I]^I \\ [\dot{\bs{\omega}}_{B_3}^I]^I
    \end{bmatrix} 
    \end{align}
    
    \item \textbf{Internal Force Vector} ($6 \times 1$):
    \begin{align}
        \mat{F}_J = 
    \begin{bmatrix}
    [\bs{f}_{J_1}]^I \\ [\bs{f}_{J_2}]^I
    \end{bmatrix}
    \end{align}
\end{itemize}

\subsection{Generalized Mass Matrix $\mathcal{M}$}
The $18 \times 18$ mass matrix requires all inertia tensors to be expressed in the inertial frame.
\begin{align}
 \mathcal{M} = \operatorname{blockdiag}([\mat{M}_v]^I, [\mat{M}_\omega]^I)   
\end{align}
\begin{itemize}
    \item \textbf{Translational Mass Matrix} ($9 \times 9$):
    \begin{align}
        [\mat{M}_v]^I = \operatorname{blockdiag}(m_{B_1}\mat{I}_3, m_{B_2}\mat{I}_3, m_{B_3}\mat{I}_3)
    \end{align}    
    \item \textbf{Rotational Inertia Matrix} ($9 \times 9$): The body-frame inertia tensors $[\mat{I}_{B_i}]^{B_i}$ are transformed into the inertial frame.
    \begin{align}
    [\mat{M}_\omega]^I = \operatorname{blockdiag}([\mat{I}_{B_1}]^I, [\mat{I}_{B_2}]^I, [\mat{I}_{B_3}]^I)        
    \end{align}
    where each block is computed as:
    \begin{align}
    [\mat{I}_{B_i}]^I = ([T]^{B_iI})^\intercal [\mat{I}_{B_i}]^{B_i} [T]^{B_iI}        
    \end{align}
\end{itemize}

\subsection{Constraint Jacobian Matrix $\mat{J}$}
The $6 \times 18$ Jacobian components must be constructed from vectors in the inertial frame.
\begin{align}
   \mat{J} = [[\mat{J}_v]^I, [\mat{J}_\omega]^I] 
\end{align}

\begin{itemize}
    \item \textbf{Linear Part} ($6 \times 9$):
    \begin{align}
      [\mat{J}_v]^I = \begin{bmatrix}
    -\mat{I}_3 & \mat{I}_3 & \mat{0} \\
    \mat{0} & -\mat{I}_3 & \mat{I}_3
    \end{bmatrix}      
    \end{align}   
    \item \textbf{Angular Part} ($6 \times 9$):
    \begin{align}
        [\mat{J}_\omega]^I = \begin{bmatrix}
    -\mat{S}([\bs{s}_{B_1J_1}]^I) & \mat{S}([\bs{s}_{B_2J_1}]^I) & \mat{0} \\
    \mat{0} & -\mat{S}([\bs{s}_{B_2J_2}]^I) & \mat{S}([\bs{s}_{B_3J_2}]^I)
    \end{bmatrix}
    \end{align}
    where $\mat{S}([\bs{v}]^F)$ denotes the skew-symmetric matrix formed from the components of tensor $\bs{v}$ in frame $F$.
\end{itemize}

\subsection{RHS Vectors}
These vectors contain all applied and velocity-dependent terms, with each term expressed in its proper coordinate frame.
\begin{itemize}
    \item \textbf{Generalized Force Vector} ($18 \times 1$):
    \begin{align}
      \mathcal{F} =
    \begin{bmatrix}
    [\bs{f}_{B_1}]^I + m_{B_1}[\bs{g}]^I \\
    [\bs{f}_{B_2}]^I + m_{B_2}[\bs{g}]^I \\
    [\bs{f}_{B_3}]^I + m_{B_3}[\bs{g}]^I \\
    ([\mat{T}^{B_1I}])^\intercal [\bs{m}_{B_1}]^{B_1} + [\bs{m}_{J_1}]^I - \mat{S}([\bs{\omega}_{B_1}^I]^I) [\mat{I}_{B_1}]^I [\bs{\omega}_{B_1}^I]^I \\
    ([\mat{T}^{B_2I}])^\intercal [\bs{m}_{B_2}]^{B_2} - [\bs{m}_{J_1}]^I + [\bs{m}_{J_2}]^I - \mat{S}([\bs{\omega}_{B_2}^I]^I) [\mat{I}_{B_2}]^I [\bs{\omega}_{B_2}^I]^I \\
    ([\mat{T}^{B_3I}])^\intercal [\bs{m}_{B_3}]^{B_3} - [\bs{m}_{J_2}]^I - \mat{S}([\bs{\omega}_{B_3}^I]^I) [\mat{I}_{B_3}]^I [\bs{\omega}_{B_3}^I]^I
    \end{bmatrix}  
    \end{align}
    
    \item \textbf{Constraint Acceleration Vector} ($6 \times 1$):
    \begin{align}
        \bs{\gamma} =
    \begin{bmatrix}
    \bs{\gamma}_1 \\ \bs{\gamma}_2
    \end{bmatrix}
    =
    \begin{bmatrix}
    -\mat{S}([\bs{\omega}_{B_1}^I]^I) \mat{S}([\bs{\omega}_{B_1}^I]^I) [\bs{s}_{B_1 J_1}]^I + \mat{S}([\bs{\omega}_{B_2}^I]^I) \mat{S}([\bs{\omega}_{B_2}^I]^I) [\bs{s}_{B_2 J_1}]^I \\
    -\mat{S}([\bs{\omega}_{B_2}^I]^I) \mat{S}([\bs{\omega}_{B_2}^I]^I) [\bs{s}_{B_2 J_2}]^I + \mat{S}([\bs{\omega}_{B_3}^I]^I) \mat{S}([\bs{\omega}_{B_3}^I]^I) [\bs{s}_{B_3 J_2}]^I
    \end{bmatrix}
    \end{align}
\end{itemize}

\section{Analysis of the Skew-Symmetry Property}\label{app:m_dot_2c}

This appendix provides a detailed derivation of the Coriolis/gyroscopic matrix $\bs{\mathcal{C}}(\bs{\nu})$ from the system equations and analyzes the properties of the matrix $\mat{N} = D^I\mathcal{M} - 2\bs{\mathcal{C}}$, which is central to the Lyapunov stability proof.

\subsection{Deriving the Components}
The two matrices in the expression, $D^I\mathcal{M}$ and $\bs{\mathcal{C}}(\bs{\nu})$ are explicitly defined as follows.

\subsubsection{\emph{Inertial Derivative of the Mass Tensor}}

The generalized mass matrix is $\mathcal{M} = \operatorname{blockdiag}(\mat{M}_v, \mat{M}_\omega)$.
\begin{itemize}
    \item The translational mass matrix $\mat{M}_v$ is constant because the body masses are constant. Its derivative is therefore zero: $D^I\mat{M}_v = \mat{0}$.
    \item The rotational inertia second-order tensor $\mat{M}_\omega = \operatorname{blockdiag}(\mat{I}_{B_1}^I, \dots, \mat{I}_{B_N}^I)$ is time-varying because the bodies rotate, changing the orientation of their inertia tensors in the inertial frame. The rate of change is given by the Poisson equation for a rotated tensor:
    \begin{align}
        D^I\mat{I}_{B_i}^I = \bs{\Omega}_{B_i}^I\mat{I}_{B_i}^I - \mat{I}_{B_i}^I\bs{\Omega}_{B_i}^I
    \end{align}
\end{itemize}

The relationship for the time derivative of a rotated inertia tensor, $D^I\mat{I}^I = \mat{S}(\bs{\omega}^I)\mat{I}^I - \mat{I}^I\mat{S}(\bs{\omega}^I)$, is known as the transport theorem for a second-order tensor. It's derived by applying the product rule to the transformation law for the tensor, combined with the kinematic equation for a rotation matrix.

The derivation begins with the fundamental relationship between the inertia tensor in the inertial frame, $\mat{I}^I$, and the constant inertia tensor in the body-fixed frame, $\mat{I}^B$. A rotation matrix, $\mat{T}^{IB}(t)$, which transforms vectors from the body frame to the inertial frame, relates them:
\begin{align}
    \mat{I}^I(t) = \mat{T}^{IB}(t) \mat{I}^B (\mat{T}^{IB}(t))^\intercal
\end{align}
The key is that $\mat{I}^B$ is constant because the body's mass distribution does not change relative to itself, while $\mat{T}^{IB}(t)$ and therefore $\mat{I}^I(t)$ change with time as the body rotates. Now take the rotational time derivative of the transformation law. Since $\mat{I}^B$ is constant, its rotational derivative with respect to the Body frame is zero, and the product rule for matrices gives:
\begin{align}
    D^I\mat{I}^I = \frac{d}{dt}(\mat{I}^I) = \frac{d}{dt}(\mat{T}^{IB}) \mat{I}^B (\mat{T}^{IB})^\intercal + \mat{T}^{IB} \mat{I}^B \frac{d}{dt}((\mat{T}^{IB})^\intercal)
\end{align}

The rotational time derivative of a rotation tensor is related to the angular velocity tensor, $\bs{\omega}^I_B$, through the Poisson kinematic equation:
\begin{align}
    \frac{d}{dt}(\mat{T}^{IB}) = \bs{\Omega}_B^I \mat{T}^{IB}
\end{align}
The derivative of the transpose is then:
\begin{align}
    \frac{d}{dt}((\mat{T}^{IB})^\intercal) = -(\mat{T}^{IB})^\intercal \bs{\Omega}_B^I
\end{align}

Now, substituting the kinematic relationships leads to:
\begin{align}
    D^I\mat{I}^I = \bs{\Omega}_B^I \underbrace{\left(\mat{T}^{IB} \mat{I}^B (\mat{T}^{IB})^\intercal\right)}_{\text{This is } \mat{I}^I} - \underbrace{\left(\mat{T}^{IB} \mat{I}^B (\mat{T}^{IB})^\intercal\right)}_{\text{This is } \mat{I}^I} \bs{\Omega}_B^I .
\end{align}
Therefore, the rotational derivative of the mass tensor is:
\begin{align}
    D^I\mathcal{M} = 
\begin{bmatrix}
\mat{0} & \mat{0} \\
\mat{0} & D^I\mat{M}_\omega
\end{bmatrix}
\quad \text{where} \quad
D^I\mat{M}_\omega = \operatorname{blockdiag}\left( \bs{\Omega}_{B_i}^I\mat{I}_{B_i}^I - \mat{I}_{B_i}^I\bs{\Omega}_{B_i}^I \right).
\end{align}

\subsubsection{\emph{The Coriolis/Gyroscopic Tensor}}
This tensor is defined before such that it reproduces the gyroscopic torques when multiplied by the generalized velocity, i.e., $\bs{\tau}_{gyro}(\bs{\nu}) = \bs{\mathcal{C}}(\bs{\nu})\bs{\nu}$, where,
\begin{align}
    \bs{\mathcal{C}}(\bs{\nu}) = 
\begin{bmatrix}
\mat{0} & \mat{0} \\
\mat{0} & \bs{\mathcal{C}}_{\omega\omega}(\bs{\nu})
\end{bmatrix}
\quad \text{and} \quad
\bs{\mathcal{C}}_{\omega\omega}(\bs{\nu}) = \operatorname{blockdiag}\left( \bs{\Omega}_{B_1}^I\mat{I}_{B_1}^I,\ldots,\bs{\Omega}_{B_N}^I\mat{I}_{B_N}^I\right) .
\end{align}

\subsection{Checking the Skew-Symmetry Property}
Now $\mat{N}$  is constructed, and its skew-symmetric property can be checked. To examine the property, only a single diagonal block of the bottom-right quadrant suffices, $\mat{N}_{i}$.
\begin{align}
    \mat{N}_{i} &= D^I\mat{I}_{B_i}^I - 2\bs{\mathcal{C}}_{i} =  \bs{\Omega}_{B_i}^I\mat{I}_{B_i}^I - \mat{I}_{B_i}^I\bs{\Omega}_{B_i}^I  - 2 \bs{\Omega}_{B_i}^I\mat{I}_{B_i}^I  = -\bs{\Omega}_{B_i}^I\mat{I}_{B_i}^I - \mat{I}_{B_i}^I\bs{\Omega}_{B_i}^I
\end{align}
Taking the transpose yields $\mat{N}_{i}^\intercal = \mat{I}_{B_i}^I\bs{\Omega}_{B_i}^I + \bs{\Omega}_{B_i}^I\mat{I}_{B_i}^I= - \mat{N}_i$, and hence $\mat{N}$ is skew-symmetric. 

\bibliography{main}
\bibliographystyle{unsrturl}
\end{document}